\documentclass[]{spie}  

\usepackage{amsmath,amsfonts,amssymb}
\usepackage{graphicx}
\usepackage[colorlinks=true, allcolors=blue]{hyperref}
\usepackage{multiraw}

\title{Fine-tuning of sign language recognition models: a technical report}

\author[a]{Maxim Novopoltsev} 
\author[b]{Leonid Verkhovtsev}
\author[c]{Ruslan Murtazin}
\author[d]{Dmitriy Milevich}
\author[e]{Iuliia Zemtsova}

\affil[a,b,c,d,e]{SBER AI, Moscow, Russia}

\authorinfo{Further author information: Maxim Novopoltsev.: E-mail: MYNovopoltsev@sberbank.ru\\Leonid Verkhovtsev.: E-mail: LRVerkhovtsev@sberbank.ru\\Ruslan Murtazin.: E-mail: RBMurtazin@sberbank.ru\\Dmitriy Milevich.: E-mail: DEMilevich@sberbank.ru\\Iuliia Zemtsova.: E-mail: YMZemtsova@sberbank.ru}

\begin{document} 
\maketitle
\begin{abstract}
Sign Language Recognition (SLR) is an essential yet challenging task since sign language is performed with the fast and complex movement of hand gestures, body posture, and even facial expressions. 
In this work, we focused on investigating two questions: how fine-tuning on datasets from other sign languages helps improve sign recognition quality, and whether sign recognition is possible in real-time without using GPU. Three different languages datasets (American sign language WLASL, Turkish - AUTSL, Russian - RSL) have been used to validate the models. The average speed of this system has reached 3 predictions per second, which meets the requirements for the real-time scenario. This model (prototype) will benefit speech or hearing impaired people talk with other trough internet. We also investigated how the additional training of the model in another sign language affects the quality of recognition. The results show that further training of the model on the data of another sign language almost always leads to an improvement in the quality of gesture recognition. We also provide code for reproducing model training experiments, converting models to ONNX format, and inference for real-time gesture recognition.
\end{abstract}
\keywords{Sign Language Recognition, Russian Sign Language, action classification, action recognition}

\section{INTRODUCTION}
\label{sec:intro}  
 Sign language recognition (SLR) is the task of recognizing individual signs or tokens called glosses from a given segment of signing video clip. There are two types of sign language recognition system: sensor- based, and vision-based. The disadvantage of the first method is that it is expensive, requires wearing sensors to recognize gestures, and is also unstable in some environments. 
 Much research endeavored to develop high-performance SLR. But most of these systems require large computational power including GPU usage. We present SLR system on CPU, that perform about 3 predictions per second on a Apple Macbook pro16 (2021) m1 pro 16gb. Our code is available at \cite{slr_rsl_github}. Considering that existing word-level russian sign language datasets do not provide a large-scale vocabulary of signs, we firstly collect large-scale word-level signs in RSL as well as their corresponding annotations. Further we will introduce a new large-scale Russian Sign Language (RSL) video dataset, containing more than 240000 gloss samples performed by 5 signers. We select 4 signers for training and the remaining 1 signer for testing.


\section{RELATED WORKS}
\subsection{Sign Language Datasets}
Sign Language Recognition (SLR) achieves significant progress and obtained high recognition accuracy in recently years due to the development on practical deep learning architectures and the surge of computational power. 
In summary, the current publicly available datasets are constrained by one or more of the following: limited vocabulary size, short video or total duration, limited domain.
Several benchmarks have been proposed for American (WLASL, MS-ASL, How2Sign, Boston ASL LVD, ASLLVD), German (DGS Kinect 40), Chinese (Isolated SLR500, NMFs- CSL), and Turkish (AUTSL) sign languages. RSL datasets, on the other hand, are scarce. Table \ref{tab:summary-datasets} provides an overview of the large-scale isolated and continuous sign language datasets.
\begin{table}[!ht]
\caption{\label{tab:summary-datasets} Summary of  sign language datasets.}
\begin{center}
\begin{tabular}{|c|c|c|c|c|c|}
\hline
Datasets & Sign Language & Task & Duration (h)  & Vocab.glosses & Glosses \\
\hline
WLASL & American & Recognition & 14 & 2000 & 21000 \\
Boston ASLLVD & American & Recognition &-&3300&9800\\
MS-ASL & American & Recognition &-&1000&25513\\
AUTSL & Turkish & Recognition & -&226&38336\\
Phoenix 14t & German & Translation &11 &1066&76000\\
How2sign & American & Translation & 79 & 16000 &-\\
RSL & Russian & Translation & 69 & 2644 & 244101 \\

\hline
\end{tabular}
\end{center}
\end{table}


Many existing sign language datasets contain isolated signs. But most real-world use continuous sign language.
There are no russian continuous sign language datasets. Our dataset consists of 2644 signs performed by 5 different signers and 244101 isolated sign video samples in total. RSL dataset can be used both for the sign language recognition task and for the sign language translation task.


\subsection{Sign language recognition.}

The early sign language automation tasks were mainly for sign language recognition. Initially, due to technical limitations, research on sign language recognition was focused on hand-crafted features computed for hand shape and motion \cite{Farhadi2007, Tamura1988, Fillbrandt2003}. Pose \cite{Buehler2009, Camgoz2017, Cooper2011, Ong2012, Pfister2014}, face \cite{Farhadi2007, Koller2015, Nguyen2008} and mouth \cite{Antonakos2015, KollerNey2014,Koller2015} have then been widely used as part of the recognition pipelines.

For real-life communication between the hearing and the deaf people, the later emerging Continuous Sign Sentences Recognition. Koller et al. \cite{Koller2017} present a hybrid approach based on CNN-RNN-HMM. More recently 3D CNNs have been adopted due to their representation capacity for spatio-temporal data \cite{Bilge2019, Camgoz2016, Huang2015, Albanie2020, Li2020}.  There have been efforts to use sequence-to-sequence translation models for sign language translation \cite{Camgoz2018}, though this has been limited to the weather discourse of RWTH-Phoenix, and the method is limited by the size of the training set. The recent work of \cite{LiYu2020} localises signs in continuous news footage to improve an isolated sign classifier. Some authors use additional modalities like RGB-D data \cite{Sincan2020}. Two recent concurrent works \cite{Albanie2020,Li2020} showed that I3D models  significantly outperform their pose-based counterparts. 
The Video Transformer Network (VTN), originally proposed by Kozlov et al. \cite{kozlov2020lightweight}, was used for isolated sign recognition on the corpus of Flemish sign language and achieved promising results (74.7\% accuracy on 100 classes), which were mainly limited by the size of the labeled dataset. Recent work \cite{DeCoster2021} apply VTN model  with hand cropping and pose flow (VTN-PF), achieves 92.92\% accuracy on the balanced test set of AUTSL.
\cite{Jiang2021} propose a Sign Language Graph Convolution Network (SL-GCN) to model the embedded dynamics and a novel Separable Spatial Temporal Convolution Network (SSTCN) to exploit skeleton features.

\section{Using pretrained models and fine tuning}
Sign language recognition (SLR) involves extracting features from videos and classifying them. The main challenge in working with sign languages is the lack of large datasets. To address this issue, large models trained on more general data are commonly used and then fine-tuned for a specific (downstream) task. In this work, we investigate the impact of incorporating data from other sign languages on the performance of a model. 

We tested the most widely used models for the Action Recognition task - VideoSWIN Transformer\cite{Liu2021} and MViT\cite{Fan2021}, pre-trained on the Kinetics600 dataset. Then, we fine-tuned the models on other sign language datasets. In the Table \ref{tab:results-our-model}, the entry "RSL → AUTSL → WLASL" implies that we took the Kinetics pre-trained network, fine-tuned it on the RSL dataset, then on the AUTSL dataset, and finally on the WLASL dataset. 

The obtained metrics show that the use of datasets from other sign languages leads to a significant improvement in the recognition of sign gestures.

\begin{table}[!ht]
\caption{\label{tab:results-our-model} Sign language recognition accuracy(\%) of fine-tuned models on test sets.}
\begin{center}

\begin{tabular}{|c|c|c|c|c|}
\hline
Train map & Model  & TOP 1  & TOP 5 & Mean Class Acc\\
\hline
\multicolumn{5}{|c|}{WLASL}\\
\hline
WLASL & Swin tiny & 44.58 & 80.37 & 41.50 \\
RSL→ WLASL & Swin tiny & 53.54 & 85.72 & 51.03 \\
RSL → AUTSL → WLASL & Swin tiny & \textbf{58.51} & 88.36 & \textbf{56.00} \\
RSL→ WLASL & MViT small & 56.88& \textbf{88.57} & 54.55 \\
\hline
\multicolumn{5}{|c|}{AUTSL}\\
\hline
AUTSL & Swin tiny & 94.33 & 99.41 & 94.29  \\
RSL → AUTSL & Swin tiny & 95.38 & \textbf{99.65} & 95.33 \\
RSL → WLASL→ AUTSL & Swin tiny & 95.62 &  99.63 & 95.59 \\
RSL → AUTSL & MViT small & \textbf{95.72} & 99.41 & \textbf{95.74} \\

\hline
\end{tabular}
\end{center}
\end{table}
\section{Real-time inference}



After the model has been trained, it is necessary to convert it to the ONNX format to use it in real time. The system takes frames from a web camera as input and produces predicted values that are displayed on the screen. The real-time operation of the system differs from the training mode. In training mode, there is usually one gloss for each video fragment. But in inference mode, there may be one gloss, multiple gloss, or no glosses in the fragment. To avoid excessive false triggers, especially when there are no glossees on the video, we selected the confidence threshold of the neural network, averaged adjacent predictions, and selected how often to send sets of frames to the neural network for prediction.

In inference mode, the input to the neural network is not individual gestures but continuous speech, and the spoken phrase on the video is used as ground truth. Therefore, if we focused on average accuracy during training, WER (Word Error Rate) became the main metric during inference. We present the WER values for different thresholds, strides, and numbers of forecasts for averaging (Table \ref{tab:optimal-parameters}). 

\begin{table}[!ht]
\caption{\label{tab:optimal-parameters} Inference time WER with different parameters on test set.  
{Avg size} means how many predictions we averaged, stride shows after how many frames relative to the width of the window the next forecast will be and the threshold shows how confident the model should be in the forecast.}
\begin{center}
\begin{tabular}{|c|c|c|c|c|c|c|c|c|c|c|c|}
\hline
\multirow{2}{*}{avg size} & \multicolumn{11}{|c|}{stride}\\
\cline{2-12}
&0.0&0.1&0.2&0.3&0.4&0.5&0.6&0.7&0.8&0.9&1.0\\
\hline
\multicolumn{12}{|c|}{threshold=0.5}\\
\hline
1&2.21&1.25&0.889&0.762&0.714&0.726&0.726&\textbf{0.71}&0.839&0.806&0.823\\
2&1.19&1.05&0.794&0.714&0.714&\textbf{0.71}&0.79&0.774&0.855&0.79&0.814\\
3&1.08&0.921&0.726&0.742&0.726&0.762&0.742&0.794&0.839&0.79&0.823\\

\hline
\multicolumn{12}{|c|}{threshold=0.9}\\
\hline
1&\textbf{0.746}&\textbf{0.746}&0.762&0.746&0.81&0.79&0.823&0.806&0.903&0.903&0.871\\
2&\textbf{0.746}&\textbf{0.746}&\textbf{0.746}&0.794&0.778&0.823&0.823&0.839&0.903&0.871&0.915\\
3&\textbf{0.746}&\textbf{0.746}&0.774&0.806&0.80&0.841&0.839&0.825&0.919&0.871&0.887\\

\hline
\multicolumn{12}{|c|}{threshold=0.99}\\
\hline
1&\textbf{0.81}&0.825&0.825&0.857&0.889&0.887&0.919&0.871&0.952&0.935&0.968\\
2&0.825&0.825&0.825&0.857&0.889&0.887&0.935&0.952&0.952&0.952&0.949\\
3&0.825&0.825&0.825&0.855&0.919&0.952&0.919&0.937&0.984&0.903&0.984\\
\hline
\end{tabular}
\end{center}
\end{table}


\section{CONCLUSION}

In this work, we focused on investigating two questions: how fine-tuning on datasets from other sign languages helps improve sign recognition quality, and whether sign recognition is possible in real-time without using GPU. For experiments, we used well-established architectures VideoSWIN Transformer and MViT. 

The results of the experiments show a significant improvement in sign recognition quality when models are fine-tuned on other sign languages. As seen in Table \ref{tab:extended-all-results-table}, our methods can achieve relatively high classification accuracy on WLASL and AUTSL validation subsets. We achieved sign recognition quality comparable to the SAM-SLR model, while our models can work on CPU in real-time, providing 2-3 predictions per second on an Apple Macbook Pro 16 (2021) M1 Pro 16GB, while SAM-SLR is an ensemble of 6 models.  

In this article, we provide the code for reproducing the training experiments as well as for converting the models into the ONNX format \cite{slr_rsl_github}.

\begin{table}[!ht]
\caption{\label{tab:extended-all-results-table} Results of SOTA models accuracy(\%) on AUTSL and WLASL test set.}
\begin{center}

\begin{tabular}{|c|c|c|c|c|c|c|c|c|c|c|c|c|}
\hline
\multirow{2}{*}{} & \multicolumn{2}{c|}{WLASL}&\multicolumn{2}{c|}{AUTSL} \\
\cline{2-5}
& TOP 1 & TOP 5  & Rank-1  & TOP 5\\
\hline
SAM-SLR\cite{Jiang2021} & \textbf{58.73} & - & \textbf{98.53} & \textbf{99.73} \\
Swin-RSL tiny (\textbf{our}) & 58.51 & 88.36  & 95.38 & 99.65 \\
MViT-RSL small (\textbf{our}) & 56.88 & \textbf{88.57} & 95.72 & 99.41 \\
I3D (pretraining: BSL-1K)\cite{Albanie2020}  & 46.82 & - & - & - \\
I3D\cite{Li2020}& 32.48 & 57.31 & - & - \\
VTN-PF\cite{DeCoster2021} & - & - & 92.92 & - \\
CNN+FPM+BLSTM+Attention (RGB-D)\cite{Sincan2020} & - & - & 62.03 & - \\
\hline
\end{tabular}
\end{center}
\end{table}



\bibliographystyle{abbrv} 
\bibliography{slt} 

\end{document}